\documentclass[lettersize,journal]{IEEEtran}
\usepackage{amsmath,amsfonts}
\usepackage{algorithmic}
\usepackage{algorithm}
\usepackage{array}
\usepackage[caption=false,font=normalsize,labelfont=sf,textfont=sf]{subfig}
\usepackage{textcomp}
\usepackage{stfloats}
\usepackage{url}
\usepackage{verbatim}
\usepackage{graphicx}
\usepackage{cite}
\usepackage{booktabs}
\usepackage[table]{xcolor}
\usepackage{multirow}
\usepackage{array}
\usepackage{colortbl}
\usepackage{afterpage}
\usepackage{float}
\usepackage[backref]{hyperref} 
\hypersetup{hidelinks}
\usepackage{colortbl}  
\usepackage{xcolor}
\usepackage{amsmath}
\usepackage{amssymb}
\usepackage{color}
\usepackage{indentfirst}
\usepackage{threeparttable}
\usepackage{pifont}
\usepackage{makecell}
\usepackage{arydshln}

\begin{document}

\title{AGHI-QA: A Subjective-Aligned Dataset and Metric for AI-Generated Human Images}

\author{Yunhao Li, Sijing Wu, Wei Sun, Zhichao Zhang, Yucheng Zhu, Zicheng Zhang, \\ Huiyu Duan, Xiongkuo Min,~\IEEEmembership{Member,~IEEE}, Guangtao Zhai,~\IEEEmembership{Fellow,~IEEE}

\thanks{Yunhao Li, Sijing Wu, Wei Sun, Zhichao Zhang, Yucheng Zhu, Zicheng Zhang, Huiyu Duan, Xiongkuo Min and Guangtao Zhai are with the Institute of Image Communication and Network Engineering, Shanghai Jiao Tong University, Shanghai, China (e-mail: \{lyhsjtu, wusijing, sunguwei, liquortect, zyc420, zzc1998, huiyuduan, minxiongkuo, zhaiguangtao\}@sjtu.edu.cn).}
}

\maketitle
\begin{abstract}

The rapid development of text-to-image (T2I) generation approaches has attracted extensive interest in evaluating the quality of generated images, leading to the development of various quality assessment methods for general-purpose T2I outputs. However, existing image quality assessment (IQA) methods are limited to providing global quality scores, failing to deliver fine-grained perceptual evaluations for structurally complex subjects like humans, which is a critical challenge considering the frequent anatomical and textural distortions in AI-generated human images (AGHIs). To address this gap, we introduce AGHI-QA, the first large-scale benchmark specifically designed for quality assessment of AGHIs. The dataset comprises $4,000$ images generated from $400$ carefully crafted text prompts using $10$ state-of-the-art T2I models. We conduct a systematic subjective study to collect multidimensional annotations, including perceptual quality scores, text-image correspondence scores, visible and distorted body part labels. Based on AGHI-QA, we evaluate the strengths and weaknesses of current T2I methods in generating human images from multiple dimensions. Furthermore, we propose AGHI-Assessor, a novel quality metric that integrates the large multimodal model (LMM) with domain-specific human features for precise quality prediction and identification of visible and distorted body parts in AGHIs. Extensive experimental results demonstrate that AGHI-Assessor showcases state-of-the-art performance, significantly outperforming existing IQA methods in multidimensional quality assessment and surpassing leading LMMs in detecting structural distortions in AGHIs.

\end{abstract}


\section{Introduction}

\begin{figure}[ht]
\centerline{\includegraphics[width=1\linewidth]{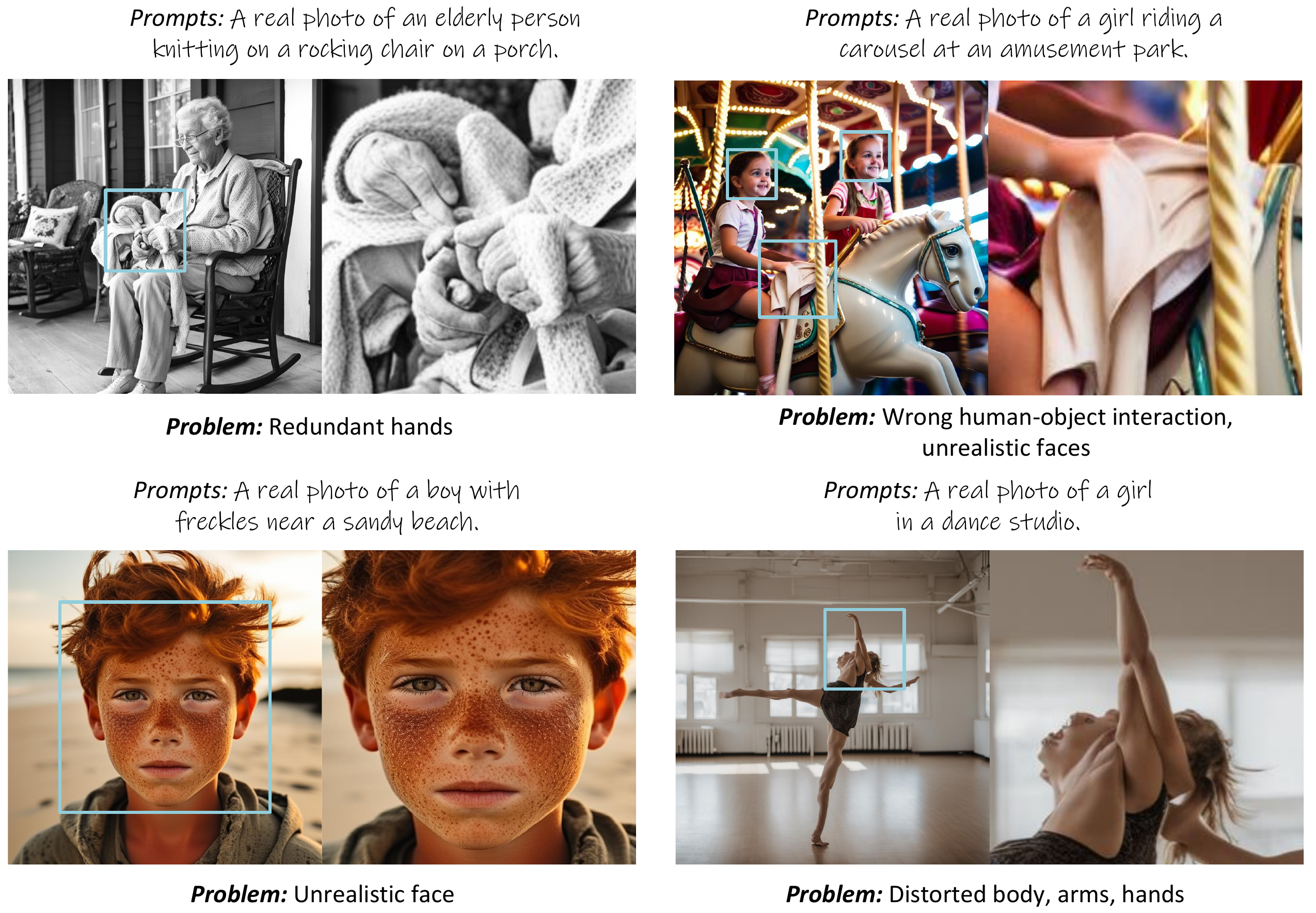}}
\caption{\textbf{Illustration} of various distortion problems in AI-generated human images from current text-to-image approaches.}
\label{human_problem}
\end{figure}

\IEEEPARstart{R}{ecent} advanced text-to-image (T2I) generation methods such as SDXL~\cite{podell2023sdxl}, Playground~\cite{li2024playground}, and Midjourney~\cite{holz2023midjourney} have demonstrated remarkable success in producing high-quality images, showing extraordinary application prospects in the fields of art creation, game production, and image editing. While current T2I models are capable of generating high-fidelity images with reasonable semantic alignment for common prompts, they still face challenges in synthesizing fully realistic and faithful content particularly when tasked with generating complex objects or human figures---often resulting in various perceptual artifacts. With the diverse distortions they often contain and the rapid proliferation of AI-generated images (AGIs), evaluating their perceptual quality has become increasingly important~\cite{xu2024imagereward, wu2023human}. Meanwhile, traditional image quality assessment (IQA) methods are primarily designed for natural images with synthetic or real-world distortions, and thus fall short when applied to AGIs. Hence, exploring and developing subjective and objective quality metrics for AI-generated images is a long-standing and significant problem.

The quality assessment of AI-generated images has emerged as a vital yet still developing field, with a growing number of studies addressing this challenge. Early works rely on metrics like Inception Score (IS) and Frechet Image Distance (FID), which evaluate the distributions of image features but fail to capture human preferences on AI-generated images. Subsequent works \cite{xu2024imagereward, kirstain2023pick, wu2023human} focus on evaluating the perceptual quality of generated images at a coarse level. AGIQA-3K \cite{li2023agiqa} further introduce a fine-grained quality assessment dataset incorporating both perceptual quality scores and text-to-image correspondence scores. More recent studies \cite{liang2024rich, wang2023aigciqa2023} construct multidimensional quality assessment datasets, including evaluations of visual quality, aesthetics, authenticity, and text-image correspondence. However, existing research primarily targets general-purpose AI-generated images encompassing objects, characters, and scenes, without specifically addressing the perceptual quality of human-centric images. Given that humans are a central subject in visual media and frequently appear in AI-generated content, assessing the quality of AI-generated human images is important and necessary. As illustrated in Fig. \ref{human_problem}, current AI-generated human images often suffer from noticeable visual artifacts such as extra or missing body parts, anatomically distorted or unrealistic figures, and overly smooth or unnatural skin textures, which significantly degrade user experience and preference.

In light of these facts, we conduct a comprehensive study on AI-generated human images and propose the first multidimensional AI-generated human image quality assessment dataset named \textbf{AGHI-QA}. AGHI-QA consists of 4000 images from 400 carefully designed text prompts covering a wide range of human-related descriptions. Each image is annotated with quality-related labels from three perspectives: a perceptual quality score, a text-image correspondence score (TI Correspondence) and identification of visible and distorted human body parts (\textit{i.e.,} face, body, arm, hand, leg, and foot). AGHI-QA serves as a valuable testbed for validating the effectiveness of existing objective quality assessment metrics on AI-generated human images. Leveraging AGHI-QA, we benchmark the performance of current T2I models and analyze their capabilities and limitations in synthesizing various types of human-centric images. Furthermore, we conduct a thorough evaluation of existing objective quality assessment metrics on AGHI-QA and assess the ability of current large multimodal models (LMMs) in detecting both visible body parts and semantically distorted body parts within AI-generated human images.

\begin{figure*}[ht]
\centerline{\includegraphics[width=0.99\linewidth]{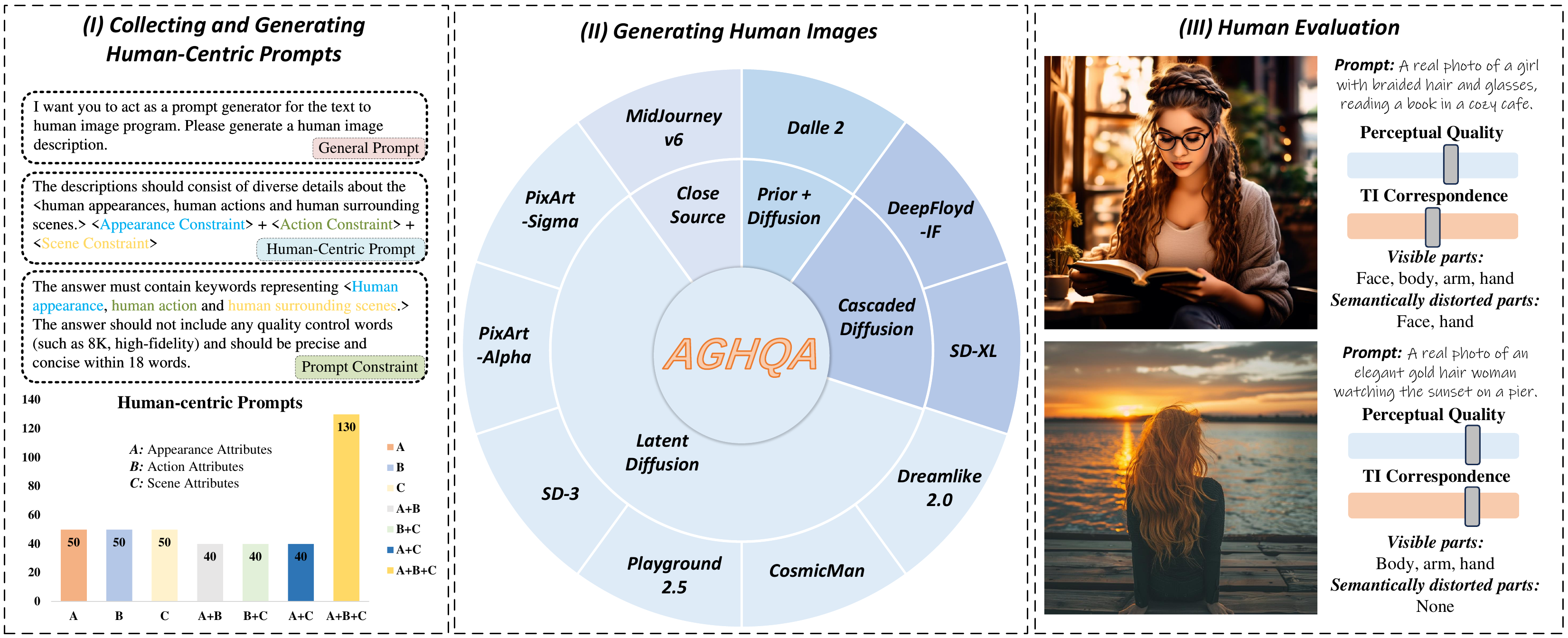}}
\caption{The overview of our quality assessment dataset \textbf{AGHI-QA}. TI-correspondence denotes the score of text-image correspondence.}
\label{teaser}
\end{figure*}

Equipped with this dataset, we propose \textbf{AGHI-Assessor}, a human-specific large multimodal model-based (LMM) image quality assessment (IQA) method for AGHIs, which provides a unified framework for regressing perceptual quality scores, evaluating text-image correspondence scores, identifying visible and semantically distorted parts of human bodies. Concretely, AGHI-Assessor consists of a visual encoder, a human-centered image cropping strategy, a holistic visual quality encoder, a text feature encoder, and a quality regressor. We adopt InternViT---the visual encoder from InternVL2.5 \cite{team2024internvl2}---as our visual quality encoder. To localize human regions, the human-centric cropping module uses a state-of-the-art human parsing network \cite{khirodkar2024sapiens} to generate cropped human-centric image patches, which are then processed alongside the original image by the visual encoder. For the holistic visual quality encoder, we utilize a pre-trained AI-generated image feature extractor \cite{kirstain2023pick} for general images to better capture complex generated artifacts. To extract dense sentence-level features from user prompts, we employ BLIP \cite{li2023blip} as our text feature encoder.  Finally, all the features are transformed to the same feature dimension of a large language model (InternLM 2.5 \cite{cai2024internlm2}) and fed into LMM to regress final scores and predict visible and distorted human body parts. The low-rank adaptation and instruct tuning techniques are adopted for effectively training our LMM-based quality assessment method. Experimental results show that our AGHI-Assessor outperforms other existing objective metrics on our AGHI-QA dataset and other general AI-Generated IQA datasets, proving its effectiveness in assessing the multidimensional quality scores of complex AI-generated human images.

The main contributions of our paper are summarized as follows:
\begin{itemize}
\item We propose the \textbf{AGHI-QA} dataset for assessing the quality of AI-generated human images. AGHI-QA consists of 4,000 images generated by 10 popular text-to-image models utilizing 400 carefully designed compositional text prompts. Each image in the dataset is annotated with an perceptual quality score, a text-image correspondence score and distorted human body part labels. All the scores are represented by mean of opinions (MOS). The diversity of text prompts and the richness quality annotations make our AGHI-QA benchmark a comprehensive and reliable testbed for evaluating the quality of AI-generated human images.

\item Based on AGHI-QA, we benchmark the performance of T2I models and analyze their strengths and weaknesses in generating different categories of human-center images. Furthermore, we propose \textbf{AGHI-Assessor}, a LMM-based objective quality assessment method, capable of automatically predicting perceptual quality scores, text-image correspondence scores, visible and semantically distorted parts of human bodies. Experimental results demonstrate that AGHI-Assessor significantly outperforms existing objective metrics on both AGHI-QA dataset and a general AI-Generated IQA datasets AGIQA-3K, highlighting its effectiveness and generalization capability.

\end{itemize}

\begin{figure*}[]
\centering
\centerline{\includegraphics[width=\linewidth]{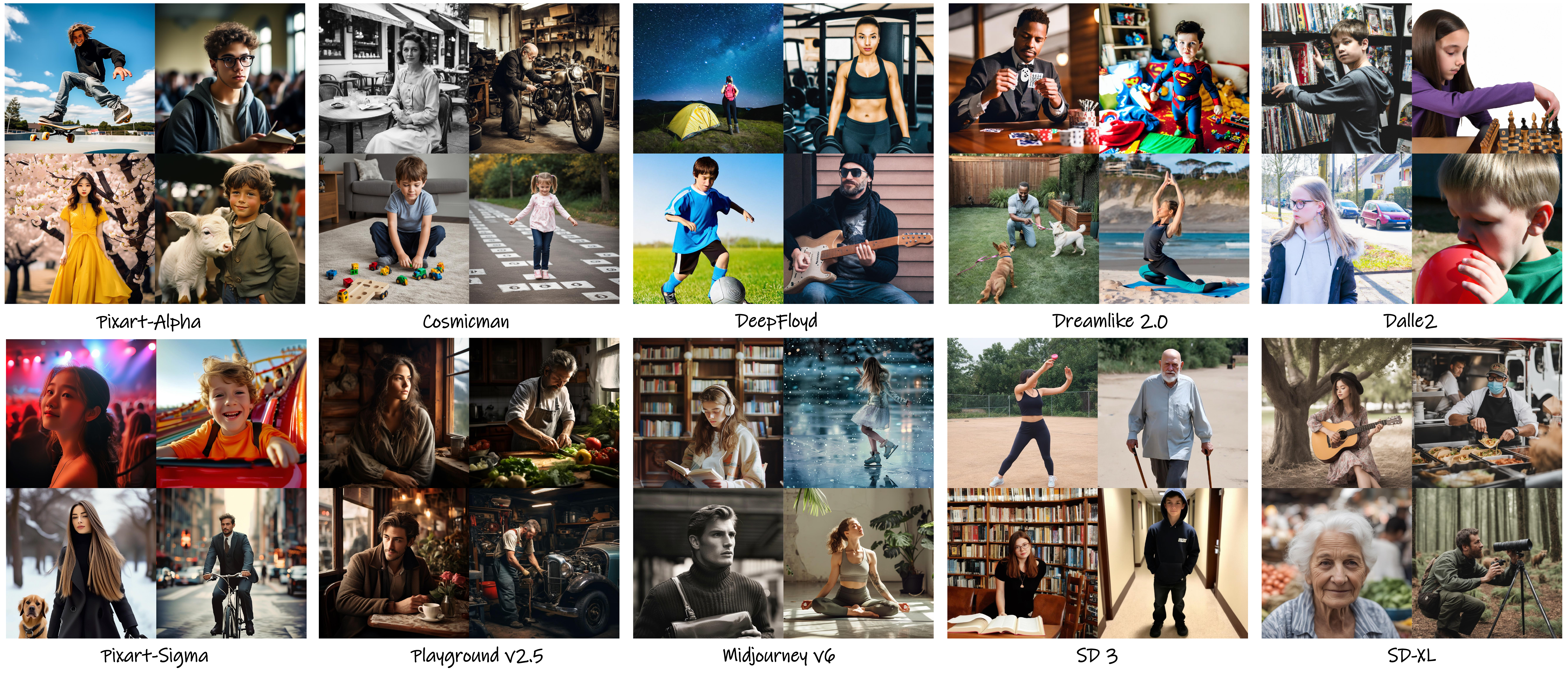}}
\caption{ \textbf{Visualization} of generated human-centric images from current popular text-to-image (T2I) methods in AGHI-QA dataset. We can observe the distinct characteristics of different T2I models. For instance, the latest models trends to generate more natural and high quality images than previous models. Playground v2.5 \cite{li2024playground} is prone to synthesize aesthetic human images in terms of color and lighting, while Cosmicman \cite{li2024cosmicman} and SD-XL \cite{dhariwal2021diffusion} are prone to generate realistic human images.}
\label{fig}
\end{figure*}

\begin{table*}[ht]
  \centering
  \caption{Comparison of \textbf{AGHI-QA} and existing AI-generated image quality assessment database \textit{(top)} and existing in-the-wild human/face quality assessment databases \textit{(bottom)}. Our AGHI-QA is the first AI-generated human image database with fine-grained annotations.}
  \renewcommand\arraystretch{1}
  \resizebox{0.92\textwidth}{!}{
    \begin{tabular}{lccccccc}
      \toprule[1pt]
      \bf Database & \bf Domain& \bf Source & \bf Models &  \bf Images& \bf Scores &\bf Ratings & \bf   Dimensions \\
      \midrule
      HPS \cite{wu2023human} &General& AIGC  &1&98807&Pair&98807&Overall Quality\\
      Pick-A-Pic ~\cite{kirstain2023pick}&General& AIGC & 3 & 500000 & Pair & 500000 & Overall Quality\\
    AGIQA-3K \cite{li2023agiqa} &General& AIGC  & 6 & 2982 & MOS & 5964 & Perceptual Quality, Alignment\\
    AIGCIQA2023 \cite{wang2023aigciqa2023} & General & AIGC & 6 & 2400 & MOS & 7200 & Perceptual Quality, Authenticity, Correspondence\\
      RichHF-18K \cite{liang2024rich}&General& AIGC & 3 & 17760 & MOS & 71040 & Plausibility, Aesthetics, Alignment, Overall Quality \\
      \hdashline
      
      CFIQA-20k \cite{su2023going} & Face& In-th-wild & - &20000 & MOS & 20000& Overall Quality  \\
      CGFIQA-40k \cite{chen2024dsl} & Face & In-th-wild  & - &  39312 & MOS & 39312 & Overall Quality \\

      FIQA \cite{liu2024assessing}  &  Face & In-th-wild & - & 42125 & MOS & 42125 & Overall Quality \\

    FaceQ \cite{liu2024f} & Face & AIGC & 14 & 4032 & MOS & 12096 & Preceptual Quality, Authenticity, Correspondence \\
      
      PIQ23 \cite{chahine2023image} & Human & In-th-wild & - & 5116 & MOS & 15348 & Overall Quality, Details, Exposure  \\

       \rowcolor{gray!20} \bf AGHI-QA &\bf Human & \bf AIGC & \textbf{10} & \bf 4000 & \bf MOS &\bf 16000 & \bf Perceptual Quality, Correspondence, Distortion Labels \\
      \bottomrule[1pt]
    \end{tabular}
    }

  \label{database}
\end{table*}

\section{Related Work}
\label{sec:related}

\subsection{Text-to-image Generation} 
Current image generation approaches can be broadly divided into three categories: GAN/VAE-based methods \cite{brock2018large, karras2019style, vahdat2020nvae}, autoregressive-based methods \cite{esser2021taming, tian2024visual, razavi2019generating} and diffusion-based methods \cite{dhariwal2021diffusion, song2020denoising, song2020score}. GAN-based methods, once popular, have become less favored due to their instability when training on large-scale text-to-image datasets. Autoregressive models \cite{sun2024autoregressive, razavi2019generating, esser2021taming} combine image tokenizers and transformers to synthesize images in a next-token prediction manner. Diffusion-based methods currently demonstrate the strongest image generation capabilities. Prior works continuously optimize the denoising learning mechanisms \cite{ho2020denoising, dhariwal2021diffusion} and sampling efficiency \cite{song2020denoising} to synthesize high-quality images. More recently, latent diffusion models \cite{rombach2022high} introduce the strategy of training diffusion models in latent space rather than pixel space, leading to significant efficiency gain. Subsequent works \cite{peebles2023scalable, bao2023all} explore scalable transformer architectures for diffusion modeling, achieving state-of-the-art performance. Modern commercial diffusion models \cite{podell2023sdxl} have been applied to various areas such as artistic creation, advertising, and marketing. Despite the impressive progress of current text-to-image models, relatively few efforts have been made toward developing human-centric generation models. Jiang \emph{et al.} \cite{jiang2022text2human} propose the first human image generation model based on VQVAE model with limited text understanding ability. CosmicMan \cite{li2024cosmicman} utilizes the stable diffusion model and optimizes it for text to human image generation.

\subsection{Image Quality Assessment}
\subsubsection{AI-generated Image Quality Assessment} 
With the rapid development of T2I models, AIGC image quality assessment has become an increasingly important research area. Compared to natural images, AIGC images may contain various new distortion types , such as overly smooth textures and anatomically unnatural image regions. Several databases \cite{wang2023aigciqa2023, liang2024rich, kirstain2023pick, li2023agiqa, wu2023human, xu2024imagereward, li2023agiqa, wang2023aigciqa2023} have been proposed to evaluate the quality of general AI-generated images. AGIQA-3K proposes the first MOS-based AI-generated image quality assessment database, and provides both perception score and alignment score. AIGCIQA2023 provides a more fine-grained quality database with scores in terms of quality, authenticity and correspondence. Recently, RichHF-18K \cite{liang2024rich} evaluates the AI-generated images with plausibility, aesthetics, alignment and overall scores. It also annotates the detailed distortion heatmaps that reflect the unrealistic regions within images. Recently, Liu \emph{et al.} \cite{liu2024f} evaluate the AI-generated face images from 3 perspectives including quality, authenticity, and correspondence. Despite these significant advances, existing datasets focus primarily on general AI-generated images, ignoring the importance of evaluating AI-generated human-centric images. Our work addresses this gap by introducing a multidimensional IQA benchmark specifically designed for AI-generated human image quality assessment.

Quality assessment methods for AI-generated images have also been widely studied in recent years. Various methods \cite{peng2024aigc, wang2024large, yang2024moe} have been proposed with specifically designed architecture such as transformer or attention module. Few work \cite{wang2024large} concentrates on the use of large multimodal models for AI-generated content quality assessment.

\subsubsection{Natural Image Quality Assessment}
Image Quality Assessment (IQA) is a long-standing research problem that aims to evaluate the quality of images under various types of distortions. Natural IQA dataset can be divided into artificially distorted image dataset and authentically distorted image dataset. Artificially distorted image datasets such as Kadid-10k \cite{lin2019kadid}, TID 2013 \cite{ponomarenko2015image}, aim at utilizing diverse human-designed distortion noises to simulate real-world distortions or distortions of specific scenes such as compression. The authentically distorted image dataset focuses on the natural distortion existing in the real world such as large in-the-wild dataset \cite{hosu2020koniq}, smart phone captured dataset SPAQ \cite{fang2020perceptual} and phone captured human image dataset \cite{chahine2023image}. In addition to the traditional general IQA dataset, many researchers also focus on the face image quality assessment (FIQA) problem, which is more related to our paper. GFIQA \cite{su2023going} introduce a in-the-wild face image database consisting of diverse distortions such as lightning, overexposure, blur etc. CGFIQA \cite{chen2024dsl} further construct a larger in-the-wild FIQA dataset. Li \emph{et al.} \cite{li2024perceptual} extend the face image quality assessment to the video area. Recently, PIQ23 \cite{chahine2023image} proposes the first traditional human image quality assessment database focusing on assessing the quality of smart phone photos. Compared to PIQ23, our AGHI-QA is the first fine-grained quality assessment dataset for AI-generated human images.

The Natural Image Quality Assessment method has also been studied for a long time. Previous works \cite{mittal2012making, mittal2011blind} utilize handcrafted features to evaluate the perceptual quality of natural images. With the development of deep neural network architecture, various works \cite{chen2024dsl, yang2022maniqa, ke2021musiq, su2020blindly, zhang2021uncertainty, zhang2018blind} seek to study different convolution network-based or transformer-based architectures. In recent years, large multimodal model-based quality assessment methods \cite{wu2023q, wu2025fvq, ge2024lmm} have also been studied for natural image or video quality assessment, demonstrating the effectiveness of large multimodal model.

\begin{table}
\setlength{\belowcaptionskip}{-0.02cm}
  \centering
\renewcommand\arraystretch{1}
\caption{An overview of selected popular text-to-image (T2I) models in AGHI-QA dataset.}
   \resizebox{\linewidth}{!}{\begin{tabular}{lcccc}
    \toprule[1pt]
     \bf Model & \bf Resolution &\bf Characteristic & \bf Open\\
    \midrule
    DeepFloyd \cite{DeepFloydIF} & 768$\times$768 & Cascaded Diffusion & \ding{51} \\
    Dalle2 \cite{ramesh2022hierarchical} & 1024$\times$1024 & Prior $\&$ Diffusion & \ding{55}  \\
    Dreamlike 2.0 \cite{dreamlike} & 1024$\times$1024 & Latent Diffusion & \ding{51} \\
    Pixart \cite{chen2023pixart} & 1024$\times$1024 & Transformer, Latent Diffusion & \ding{51} \\
    Pixart-Sigma \cite{chen2025pixart} & 1024$\times$1024 & Transformer, Latent Diffusion & \ding{51} \\
    Playground v2.5 \cite{li2024playground} & 1024$\times$1024 & Aesthetics, Latent Diffusion & \ding{51} \\
    SD XL \cite{podell2023sdxl} & 1024$\times$1024 & Cascaded Diffusion & \ding{51} \\
    CosmicMan \cite{li2024cosmicman} & 1024$\times$1024 & Latent Diffusion & \ding{51} \\
    Midjopurney v6 \cite{holz2023midjourney} & 1024$\times$1024 & Diffusion & \ding{55} \\
    SD 3 \cite{esser2024scaling} & 1024$\times$1024 & Transformer, Latent Diffusion & \ding{51} \\

    \bottomrule[1pt]
  \end{tabular}}
  \label{human_model}
\end{table}

\section{AGHI-QA Dataset Construction and Analysis}

In this section, we will first describe the construction of the AGHI-QA dataset and then provide a comprehensive analysis.

\subsection{Data preparation}

\subsubsection{Prompts Selection} 
To thoroughly evaluate the performance of current T2I models, it is essential to include diverse human-centric text prompts for generating human images. To achieve this, we carefully design a rule-based approach for constructing complex prompts by selectively assembling desired attributes using GPT-4o. Specifically, we first define three fundamental elements that comprehensively describe a human in a single image: appearance (A), action (B) and scene (C). These elements serve as the core attributes for prompt construction. We consider seven combination modes---A, B, C, A+B, A+C, B+C, and A+B+C---with each mode comprising 50, 50, 50, 40, 40, 40, and 120 prompts, respectively. We generate more prompts for the A+B+C combination (appearance + action + scene) to assess model performance on complex, information-rich inputs Subsequently, we utilize GPT-4o to construct precise, high-quality sentences to serve as text prompts for image generation. For each specific attribute, we manually adjust the instructions to make sure that GPT-4o can precisely generate diverse sentences that integrate these attributes. For instance, we utilize the following template prompt to generate abundant text prompts for the A+B+C combination (appearance + action + scene):

\begin{quote}
\#  \textit{I want you to act as a prompt generator for the text to human image program. Please generate 100 human image descriptions. The descriptions should consist of diverse details about the human appearances, human actions and human surrounding scenes. The human appearances may include wearings (such as clothes, pants, hats, shoes), hair attributes, face appearance or figures. The human actions should be actions in the real world. The human surrounding scenes should be scenes in the real world. The answer must contain keywords representing human appearance, human action and human surrounding scenes. The answer should not include any quality control words (such as 8K, high-fidelity) and should be precise and concise within 18 words.} \#
\end{quote}

As for the other combinations (A, B, C, A+B, A+C, B+C), we slightly modify the template prompt by removing the corresponding requirements related to appearance, action or scene. For each attribute combination, we first generate numerous prompts and conduct feature clustering based on the extracted Bert \cite{devlin2019bert} features from all the prompt sentences, making our final prompts covering a wide range of human-related texts. We select one prompt from each clustered prompt set and then manually assess the appropriateness of each prompt. If the prompt is considered unsafe or negative, we replace the prompt with another prompt from the same clustered texts. Finally, considering our goal is to evaluate the plausibility and quality of real human images instead of cartoon character images, hence a "a real photo of" constraint is adopted to every prompt. In this way, we create 400 unique textual prompts that depict a range of human images.

\subsubsection{Text-to-Image Model Selection}

Considering the rareness of specific human image generation models \cite{li2024cosmicman}, we select 9 representative general T2I models and 1 human foundation T2I model \cite{li2024cosmicman} to construct our AGHI-QA dataset. As shown in Fig. \ref{teaser}, AGHI-QA includes 9 general T2I models (DeepFloyd \cite{DeepFloydIF}, Dalle 2 \cite{ramesh2022hierarchical}, Dreamlike 2.0 \cite{dreamlike}, Pixart \cite{chen2023pixart}, Pixart-Sigma \cite{chen2025pixart}, SD-XL \cite{podell2023sdxl}, SD 3 \cite{esser2024scaling}, Playground v2.5 \cite{li2024playground}, Midjourney v6 \cite{holz2023midjourney}) and 1 human foundation T2I model CosmicMan \cite{li2024cosmicman}. All the prompts are fed into these models to synthesize corresponding images. For open-sourced models, we directly utilize the publicly available model weights. For closed-source models, we utilize their official APIs to synthesize human images.

\subsection{Quality Assessment Dimensions}

To comprehensively evaluate the quality of generated human images, we annotate the human images from two perspectives.

\subsubsection{Visual Quality Scoring}

We first select two key dimensions: \emph{perceptual quality} and \emph{text-image correspondence}. Subjects are asked to give continuous scores for each dimension. Concretely, for perceptual quality, subjects assess various aspects including the naturalness of the generated human structure, the presence of technical distortions, and overall aesthetic appeal. For text-image correspondence score, subjects evaluate how accurately the image content reflects the details, human subjects, and actions described in the generation prompt. During the experiment, scores of both dimensions are constrained from 0 to 5, where 0 indicates the lowest quality and 5 indicates the highest.

\subsubsection{Human Body Part Distortion Identification} Beyond merely annotating continuous scores, we also provide detailed distortion labels of human images. Hence, we carefully annotated the visibility and whether each body part (face, body, arm, hand, leg, foot) is distorted. In total, we collect 12 binary labels for each AI-generated human image to support fine-grained semantic analysis of human body part distortions.

\subsection{Subjective Experiment}

In this section, we introduce the experiment setup and the dataset processing method in detail.

\subsubsection{Participants and Experiment Setup}

To obtain the MOS for each AI-generated human image, we invite a total of 23 subjects to rate the perceptual quality score and text-image correspondence score. For perceptual quality, subjects are asked to score according to the naturalness of generated human structure, the technical distortion, and the aesthetic quality. For text-image correspondence quality, subjects are asked to score according to the degree of text-image matching. The subjects utilize two sliders from 0 to 5 to separately rate the perceptual quality and text-image correspondence. After completing the scoring task, we invite 3 subjects to annotate the human distortion types. Concretely, the subjects are asked to list all the visible body parts from face, body, arm, hand, leg, and foot in each image and provide binary labels that describe whether each body part is distorted.

We conduct the subjective experiment under the guidance of ITU-R BT.500-14 \cite{series2012methodology}. The experimental environment is designed to simulate a typical indoor home setting with standard lighting conditions. The AI-generated human images, along with their corresponding prompts, are displayed on a 1920$\times$1080 resolution screen. 23 participants (13 males and 10 females) with normal or corrected vision are recruited in our experiments. Before starting rating, each individual received thorough instructions on the evaluation process. The formal experiment was split into 10 sessions per subject, with each session containing a randomly selected subset of the AGHI-QA dataset. Each session is seted to last approximately 1 hour, then the participant is asked to rest for 10 minutes every 30 minutes.

\begin{figure*}[ht]
\centerline{\includegraphics[width=0.98\linewidth]{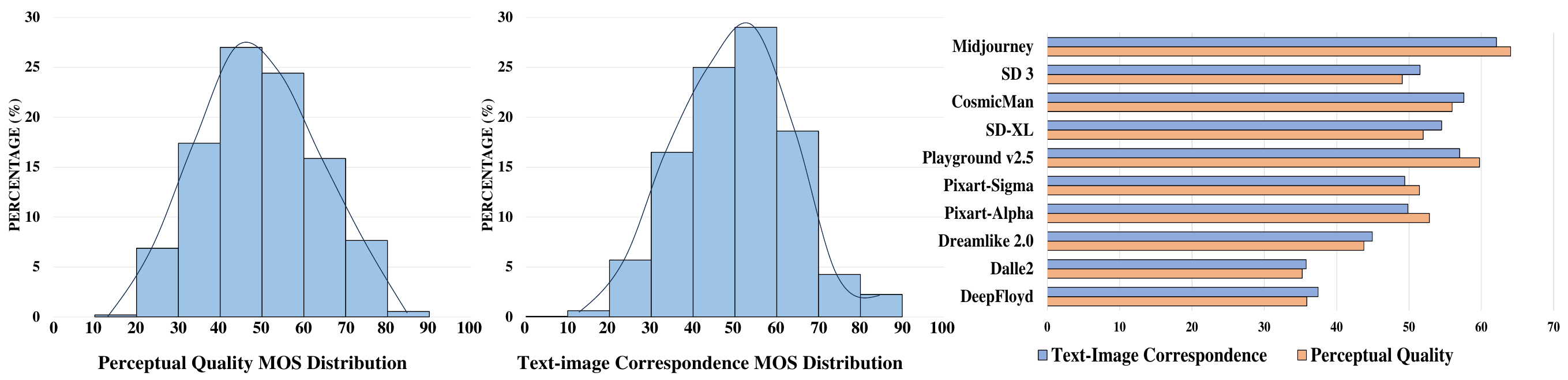}}
\caption{\textbf{Visualization} of MOS distribution and model-wise comparison results for \textbf{AGHI-QA} dataset. }
\label{mos_dis}
\end{figure*}

\subsubsection{Subjective Data Processing}

After collecting the raw scores from each subject, we follow the ITU-R BT.500-14 \cite{series2012methodology} guidelines to process our subjective data, focusing on identifying outliers and disqualifying unreliable subjects. For each assessment dimension, we first analyze the kurtosis of the raw scores assigned to each image to determine whether the distribution is Gaussian or non-Gaussian. For scores following a Gaussian distribution, any value deviating more than two standard deviations (std) from the mean is identified as an outlier. For non-Gaussian distributions, we identify the outlier score if it exceeded $\sqrt{20}$ standard deviations from the mean score. If a participant's outlier rate exceeds 5\% in any evaluation dimension, all of their scores are removed. In our experiment, no individuals are rejected, and the overall outlier rejection rate remains at approximately $2\%$ for all evaluations. After filtering, we transform the valid raw scores into Z-normalized mean opinion score (MOS) scores, which are in a range of 0 to 100. The final MOS is calculated as follows:

$$z_i{}_j=\frac{r_i{}_j-\mu_i{}_j}{\sigma_i},\quad z_{ij}'=\frac{100(z_{ij}+3)}{6},$$
$$\mu_i=\frac{1}{N_i}\sum_{j=1}^{N_i}r_i{}_j, ~~ \sigma_i=\sqrt{\frac{1}{N_i-1}\sum_{j=1}^{N_i}{(r_i{}_j-\mu_i{}_j)^2}},$$ 
where $r_{ij}$ denotes the raw ratings annotated by the $i$-th subject according to the $j$-th video. $N_i$ is the number of videos judged by subject $i$. Finally, the MOS of the $j$-th video is calculated by averaging the rescaled z-scores through the following function: $$MOS_j=\frac{1}{M}\sum_{i=1}^{M}z_{ij}' ,$$ where $MOS_j$ denotes the MOS for the $j$-th AI-generated human image, $M$ is the number of subjects, and $z'_i{}_j$ are the rescaled Z-score.

\begin{figure*}[ht]
\centerline{\includegraphics[width=\linewidth]{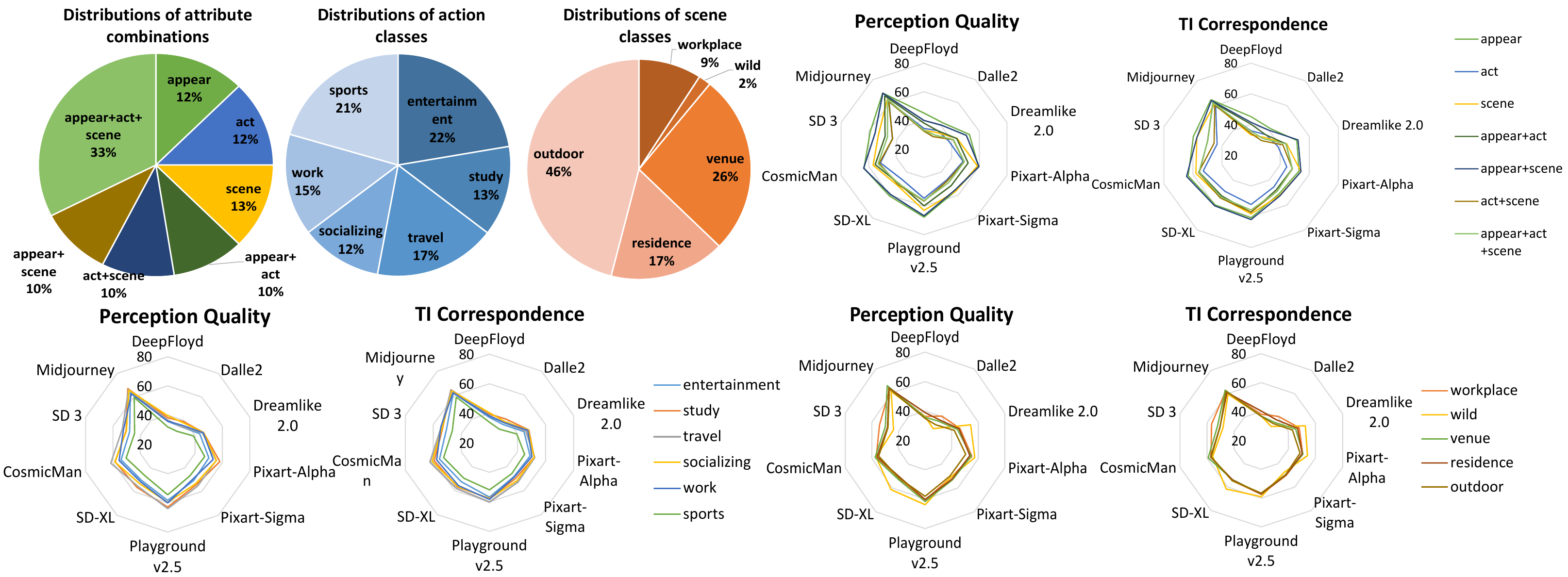}}
\caption{\textbf{Dataset Distribution Analyze.} The top left corner shows the distributions of text prompts under different categories. The rest area shows the detailed distributions of MOS scores across different combinations of attributes, different action classes and different scenarios.}
\label{dataset_distribution}
\end{figure*}

\begin{figure}[ht]
\centerline{\includegraphics[width=\linewidth]{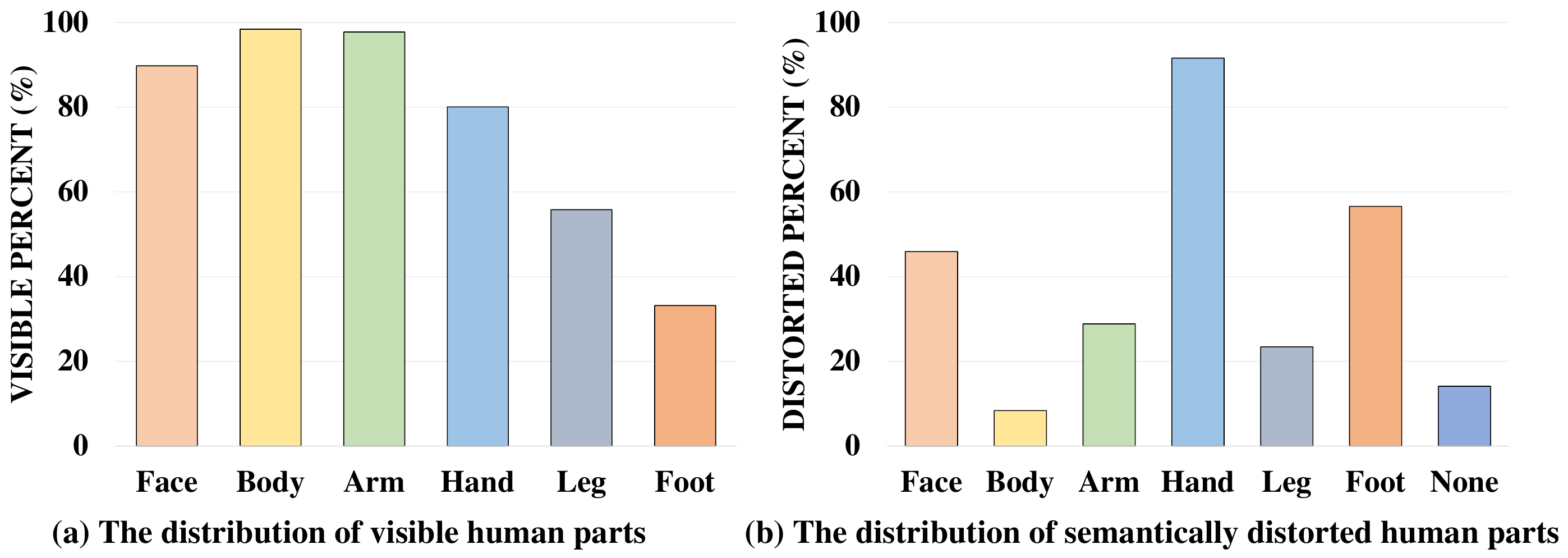}}
\caption{The distribution of the semantical distortion for each human part. The histogram (a) shows the percentage of human body parts visible in all AI-generated human images. The histogram (b) shows the percentage of semantical distorted human parts in all AI-generated human images.}
\label{part_dis}
\end{figure}

\subsection{Dataset Statistics and Analysis}

In this section, we conduct a comprehensive analysis of the AGHI-QA dataset. We analyze the overall MOS distribution, model-wise performance comparison, prompt-wise performance comparison, and human body part distortion distribution.

\subsubsection{Perceptual Quality v.s. Text-Image Corresondence} The overall MOS score distribution is shown in Fig. \ref{mos_dis}. We can observe that both perceptual quality scores and text-image correspondence scores cover a wide distribution from 0 to 100. We can also observe that the correspondence scores are integrally higher than perceptual quality scores, demonstrating that current T2I models are more effective at capturing human-related textual prompts than at generating physically plausible human figures. The reason may be that current T2I models can reliably handle the layout of the image and basic concepts from text descriptions, whereas generating anatomically correct and visually coherent human figures remains a significant challenge for current T2I models.

\subsubsection{Model-wise Comparison} Fig. \ref{mos_dis} also summarizes the results of model-wise comparison. First, we can observe that among all the models, Midjourney v6 achieves the best performance in terms of both perceptual score and text-image correspondence score, demonstrating the unique advantage of closed-sourced T2I models. Among all the open-sourced T2I models, Playground v2.5 \cite{li2024playground} and Cosmicman \cite{li2024cosmicman} stand out as the top performers across both dimensions. Playground v2.5 achieves the highest perceptual quality score, which is remarkably close to Midjourney v6, likely due to its training on a large, carefully curated aesthetic image dataset. On the other hand, CosmicMan achieves the highest correspondence score, benefiting from fine-tuning on a high-resolution real human image dataset. In contrast, Dalle 2 and DeepFloyd rank lowest in both dimensions, reflecting their performance lag relative to more advanced models. Moreover, we can notice that SD-XL is slightly better than Stable Diffusion 3, likely due to its coarse-to-fine generation strategy and the inclusion of an image refinement stage.

\subsubsection{Prompt-wise Comparison}

We further analyze MOS distributions across different prompt categories, as shown in Fig. \ref{dataset_distribution}. We specifically divide the text prompts into three types including different combinations of attributes, different action classes, and different scenes. We can first observe that AGHI-QA contains diverse and sufficient number of prompts under each category. Then for different combinations of attributes, as shown in the first row of Fig. \ref{dataset_distribution}, we can observe that ``appearance" and ``appearance+scene" combinations achieve the highest 2 scores in terms of both perceptual quality and text-image correspondence. It is mainly because T2I models can easily synthesize the general human appearance in a static scene. We also find that T2I models are prone to generate humans with normal poses or half-length photos, making it easier to synthesize human images. In addition, we can see that ``action" and ``action+scene" combinations achieve the worst 2 scores on both dimensions. The reason can be attributed to that T2I models are still struggling on generating action words. We can also see that Playground v2.5 demonstrates excellent performance on generating human images which do not contain obvious actions, but it struggles on action-related human images comparing to Midjourney. For different action classes, we concretely divide the actions happened in our dataset into 6 classes (entertainment, study, travel, socializing, work, sports) and report the corresponding MOS performance. "Sports" class undoubtedly gets the lowest average MOS, showing the limitation of synthesizing action and human-object interaction for current T2I models. "Study" and "travel" achieve the best 2 average MOS on both 2 dimensions. For different scenarios, we concretely divide the scenes occurred in our dataset into 5 classes (workplace, wild, venue, residence, outdoor) and report their corresponding MOS performance. Every category achieves similar performance in terms of both dimensions.

\subsubsection{Semantic Distortion Identification Analyze}

We analyze all visible body parts present in the 4000 AI-generated human images from the AGHI-QA dataset, along with the distortions identified for each part, as illustrated in Fig. \ref{part_dis}. From Fig. \ref{part_dis} (a), it is clear that the face, body, arm, and hand appear most frequently in all AI-generated human images, whereas the leg and foot appear far less frequently. Concretely, we find that human foot is often neglected by current T2I models and current models are prone to generate upper body shots. As for Fig. \ref{part_dis} (b), the human body part shows a few distortions among all the dataset, while the hand and foot introduce a large proportion of distortions. The reason is that the hand is very complex because it has five fingers which can be treated as rigid parts. And foot is usually uncommon in real world image dataset, introducing difficulty on synthesizing foot.

\begin{figure*}[ht]
\centerline{\includegraphics[width=\linewidth]{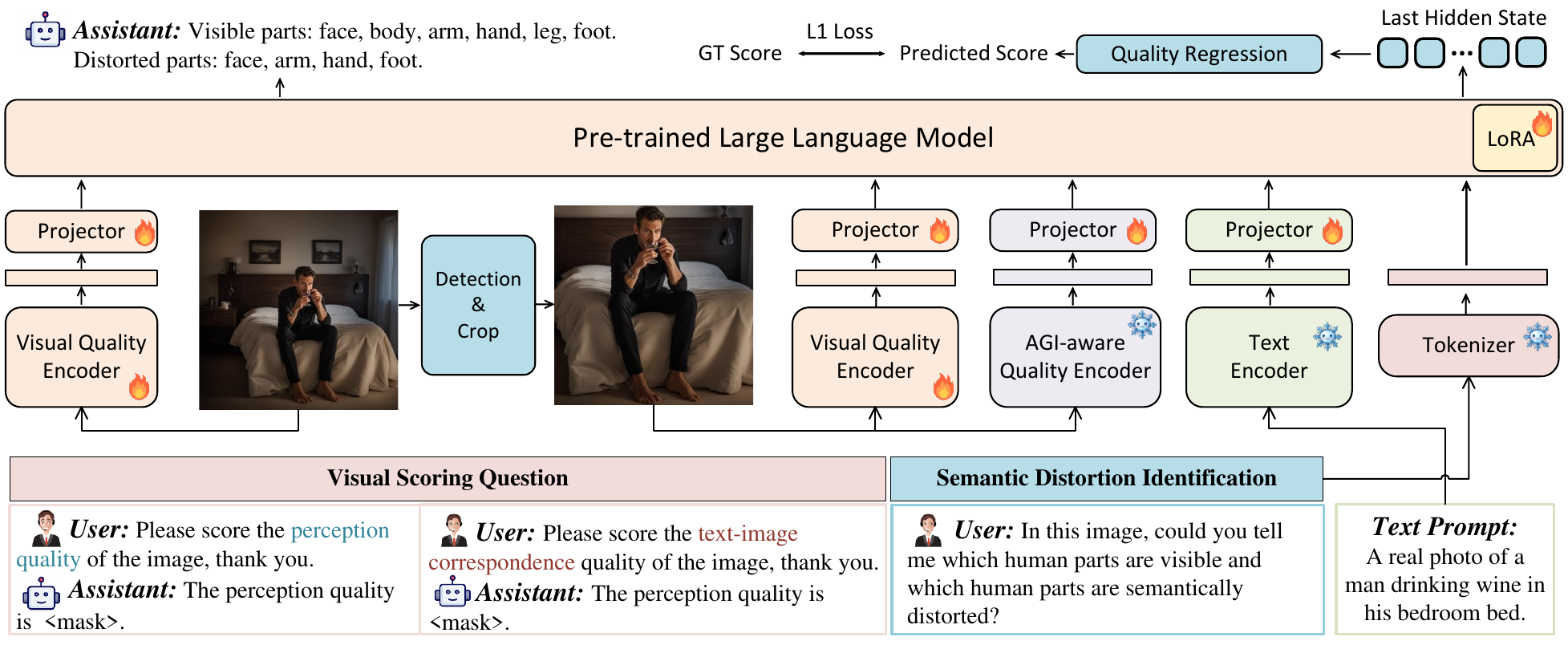} }
\caption{\textbf{Overview of our proposed AGHI-Assessor.} Our framework consists of a visual quality encoder to extract visual features of the original image and the cropped human-centered image, a AGI-aware quality encoder to extract quality features of AI-generated image and a text encoder to extract text-related features.}
\label{fig_method}
\end{figure*}


\section{The Proposed Method}

In this section, we provide a detailed introduction of our proposed LMM-based AGHI-Assessor to specifically evaluate AI-generated human images, as demonstrated in Fig. \ref{fig_method}.

\subsection{Overall Architecture} The AGHI-Assessor takes both AI-generated human images and user's text prompts as input, then regress the perceptual quality score, text-image correspondence score, and generates an answering sentence that identifies the visible and distorted human body parts. AGHI-Assessor first extracts visual features and AIGC-related features from original AI-generated human images, then utilizes a popular human parsing method to locate the human position and obtains cropped human-centric images for extracting human-centric visual features. It also utilizes a powerful text encoder to obtain text features from user prompts. Finally, all extracted feature tokens are concatenated together and fed into a large language model to regress the final scores and generate answers.

\subsection{Model Design}

\subsubsection{Visual Quality Encoder} We select a powerful pretrained vision transformer InternViT \cite{chen2024internvl} to serve as our visual quality encoder. InternViT is a large-scale vision encoder that is trained on the LAION-en dataset \cite{schuhmann2022laion} using text-to-image contrastive learning, allowing it to capture the full content of an image for better quality-aware feature extraction. In our framework, we utilize InternVit to extract features from both the original generated human images and the cropped human-centric images which will be introduced in the following.

Moreover, to ensure the compatibility between the extracted visual features and the latent dimension of the large language model (LLM), we employ a two MLP layers as a projector layer to transform the visual features. This transformation process is formally expressed as

\begin{equation}
f_{v} = E_{I}(x), 
\end{equation}
where $f_{v}$ denotes the extracted typical visual quality feature, $E_{I}$ denotes the InternViT encoder.

\subsubsection{Human-Centered Image Cropping} Considering the importance of human in our collected human-centric image dataset, simply feeding the visual features of the original images into the LMM model may lose important human-related information. Hence, we utilize the state-of-the-art human segmentation model sapiens \cite{khirodkar2024sapiens} to locate the human in every image and specifically crop a square image $x_{h}$ where the human in perfectly located in the center of the cropped image. Then we utilize visual quality encoder to obtain the human-centered feature:

\begin{equation}
x_{h} = Crop(x),
f_{hc} = E_{i}(x_{h}),
\end{equation}

where $f_{hc}$ denotes the extracted human-centered feature.

\subsubsection{AGI-aware Quality Encoder} To extract robust quality representations for AI-generated images (AGIs), we utilize a vision transformer model pre-trained on the Pick-a-Pic \cite{kirstain2023pick} dataset which is a comprehensive general text-to-image quality assessment benchmark to capture the global quality attributes of generated images. For each AI-generated human image, this quality-optimized ViT encoder generates holistic quality embeddings:

\begin{equation}
f_{hv} = ViT(x),
\end{equation}
where $f_{hv}$ denotes the holistic visual quality feature of image $x$.

\subsubsection{Text Feature Encoder} Considering that text-to-image methods aim to generate plausible images that align with text prompts, it is significant to evaluate the correspondence between human images and text prompts. To this end, we use the BLIP \cite{li2023blip} text encoder to obtain the relevant features $f_t$ of the text prompt.

\subsubsection{Feature Projection and Quality Regression} After obtaining various features $f_{v}$, $f_{hc}$, $f_{hv}$, and $f_{t}$, they are all projected into a shared feature space through different MLP layers to align with the tokens of the text questions, as demonstrated in Fig. \ref{fig_method}. Concretely, we apply three projection modules which can map the visual features, holistic visual features, and text features into the language space. Regarding the output of quality scores, different from previous research \cite{wu2023q}, which transforms predicted MOS scores into 5 discrete text-defined levels (bad, poor, fair, good and excellent), we utilize an MLP network to directly regress numericial scores of the last hidden states of the LMM model. By predicting numerical scores instead of discrete levels, the AGHI-Assessor is able to understand subjective perception of humans more precisely.

\subsection{Model Training Strategy}

\subsubsection{Instruct Tuning} Utilizing the generalization ability of the LLM model, our AGHI-Assessor model can be suitable for visual scoring and QA tasks. As demonstrated in Fig. \ref{fig_method}, we create diverse question-answer prompts containing both multidimensional visual scoring QA pairs and distortion parts identification QA pairs. For different QA instructions, we concretely train feature projectors to align the text-related features and the visual features for joint reasoning. 

\subsubsection{LoRA Adaptation} Fully fine-tuning LLMs is useful to improve model performance, but it is considered resource-consuming. Hence, we adopt LoRA which is an effective fine-tuning technique. We concretely apply LoRA on both visual encoder and llm. Suppose a model layer weight matrix is denoted by $W\in R^{d\times k}$, LoRA adds a small weight change $\bigtriangleup W$ on it and $\bigtriangleup W = BA$, where $B\in R^{d\times r}$, $A\in R^{r\times k}$. $r$ is considered as a rank parameter, which is usually a small number. Considering the original output process as $h = Wx$, the detailed LoRA can be:
\begin{equation}
h = Wx + \bigtriangleup Wx = (W + BA)x.
\end{equation}

\subsubsection{Training Loss} Considering the quality regressor can directly produce a contiguous score, we utilize a simple L1 loss to supervise the predicted scores and gt scores, $loss = |\hat{mos} - mos|$, where $\hat{mos}$ and $mos$ are the predicted scores and ground truth scores. For visible parts and semantically distorted parts identification tasks, we construct question answer pairs according to the ground truth labels and utilize cross entropy loss to supervise the training of LMM.


\begin{table*}[t]\scriptsize
\setlength{\belowcaptionskip}{-0.01cm}
\belowrulesep=0pt
\aboverulesep=0pt
\renewcommand\arraystretch{1.1}
  \caption{{\bf Performance comparisons on the AGHI-QA.} All the models are trained on AGHI-QA using the scores from each perspective. $\clubsuit$, $\spadesuit$, $\diamondsuit$ and $\heartsuit$ denote the traditional no-reference (NR) IQA method,  vision language pre-training method, deep learning-based IQA method, and deep learning-based AI-generated IQA method.}
  \centering
 \begin{tabular}{lp{1.2cm}<{\centering}p{1cm}<{\centering}p{1cm}<{\centering}p{1cm}<{\centering}p{1cm}<{\centering}p{1cm}<{\centering}p{1cm}<{\centering}p{1cm}<{\centering}p{1cm}<{\centering}p{1cm}<{\centering} }
    \toprule[1pt]
     \multirow{2}{*}{\bf Methods} & \multicolumn{3}{c}{\bf Perceptual Quality} & \multicolumn{3}{c}{\bf Text-Image Correspondence} & \multicolumn{3}{c}{\bf Average}\\
    \cmidrule(lr){2-4} \cmidrule(lr){5-7} \cmidrule(lr){8-10}
    & SRCC$\uparrow$& PLCC$\uparrow$ & KRCC$\uparrow$& SRCC$\uparrow$& PLCC$\uparrow$ & KRCC$\uparrow$& SRCC$\uparrow$& PLCC$\uparrow$ & KRCC$\uparrow$\\
    \midrule
    $\clubsuit$BRISQUE \cite{mittal2011blind} & 0.4193 & 0.3744 & 0.3519 & 0.4063 & 0.3691 & 0.3428 &  0.4128 & 0.3718 & 0.3474 \\
    $\clubsuit$NIQE \cite{mittal2012making} & 0.3317 & 0.2852 & 0.3048 & 0.2657 & 0.2493 & 0.2543 & 0.2987 & 0.2672 & 0.2796 \\
     
     \hdashline  
    $\spadesuit$CLIP-Score \cite{radford2021learning} & 0.0179 & 0.0105 & 0.0119 & 0.0491 & 0.0461 & 0.0327 & 0.0335 & 0.0283 & 0.0223 \\

    $\spadesuit$BLIP-Score \cite{li2023blip} & 0.1469 & 0.1461 & 0.0981 & 0.2324 & 0.2485 & 0.1561 & 0.1897 & 0.1973 & 0.1271 \\
    $\spadesuit$Aesthetic Score \cite{schuhmann2022laion} & 0.2818 & 0.2913 & 0.1903 & 0.2849 & 0.3114 & 0.1917 & 0.2833 & 0.3014 & 0.1910 \\

     $\spadesuit$PickScore \cite{kirstain2023pick} & 0.4774 & 0.4778 & 0.3285 & 0.4922 & 0.5029 & 0.3392 & 0.4849 & 0.4904 & 0.3339 \\

     \hdashline 
     $\diamondsuit$DBCNN \cite{zhang2018blind} &  0.7294 & 0.7340 & 0.5361 & 0.6623 & 0.6681 & 0.4758 & 0.6958 & 0.7011 & 0.5095   \\

     $\diamondsuit$HyperIQA \cite{su2020blindly}& 0.7367 & 0.7445 & 0.5418 & 0.6673 & 0.6738 & 0.4761 & 0.7020 & 0.7092 & 0.5089 \\

     $\diamondsuit$UNIQUE \cite{zhang2021uncertainty} & 0.7681 & 0.7694 & 0.5774  & 
     0.6845 & 0.6927 & 0.4988 & 0.7263 & 0.7311 & 0.5381 \\
     
     $\diamondsuit$MUSIQ \cite{ke2021musiq}& 0.7755 & 0.7816 & 0.5824 & 0.6971 & 0.7005 & 0.5087 & 0.7363 & 0.7410 & 0.5455 \\

     $\diamondsuit$DN-PIQA \cite{sun2024dual} & 0.8014 & 0.8033  & 0.6071 &  0.7328 & 0.7369 &  0.5265   & 0.7671  & 0.7707  & 0.5668  \\

     $\diamondsuit$MANIQA \cite{yang2022maniqa} & 0.8109 & 0.8167 & 0.6180 & 0.7414 & 0.7411 & 0.5445 & 0.7762 & 0.7789 & 0.5813 \\

     $\diamondsuit$DSL-FIQA \cite{chen2024dsl} & 0.8210 & 0.8242 & 0.6276 & 0.7588 & 0.7563 & 0.5613  & 0.7899 & 0.7903 & 0.5944 \\

     \hdashline 
     $\heartsuit$IPCE \cite{peng2024aigc} & 0.8078 & 0.8157 & 0.6179 & 0.7591 & 0.7652 & 0.5661 & 0.7834 & 0.7904 & 0.5920 \\

     $\heartsuit$MA-AGIQA \cite{wang2024large} & 0.8007 & 0.8099 & 0.6052 & 0.7274 & 0.7378 & 0.5328 & 0.7641 & 0.7738  & 0.5690 \\
     
    \rowcolor{gray!20} AGHI-Assessor (Ours) &  \bf{0.8447} & \bf{0.8504} & \bf{0.6625} & \bf{0.7818} & \bf{0.7864} & \bf{0.5907} &  \bf{0.8132} & \bf{0.8161} & \bf{0.6266} \\

    \bottomrule[1pt]
  \end{tabular}
  \label{benchmark_res}
\end{table*}


\section{Experiments}

\begin{table*}[t]\scriptsize
\setlength{\belowcaptionskip}{-0.01cm}
\belowrulesep=0pt
\aboverulesep=0pt
\renewcommand\arraystretch{1.1}
  \caption{\textbf{Evaluation} of LMM models on identifying the visible parts and semantically distorted parts of human body in AGHI-QA dataset.}
  \centering
 \begin{tabular}{lp{1.4cm}<{\centering}p{1.4cm}<{\centering}p{1.4cm}<{\centering}p{1.4cm}<{\centering}p{1.4cm}<{\centering}p{1.4cm}<{\centering}p{1.4cm}<{\centering} }
    \toprule[1pt]
     \multirow{2}{*}{\bf Methods} &\multicolumn{7}{c}{\bf Occurrences / Distortions} \\
    \cmidrule(lr){2-8} 
    & Face & Body & Arm & Hand & Leg & Foot & Average \\
    \midrule

     Deepseek VL-7B \cite{lu2024deepseek} &  0.570/0.495 &  0.827/0.900  &  0.661/0.693 & 0.718/0.138 & 0.581/0.771 & 0.413/0.436  & 0.628/0.572 \\

     InternVL 2-8B \cite{team2024internvl2} &  0.731/0.510  & 0.625/0.918  & 0.843/0.702  & 0.901/0.111 &  0.785/0.762 & 0.925/0.424  &  0.802/0.575  \\

     InternVL 2.5-8B \cite{chen2024expanding} &  0.311/0.514  &  0.665/0.918 & 0.752/0.697 & 0.841/0.121  & 0.836/0.767  & 0.893/0.432  &  0.716/0.571 \\

     Qwen VL 2.5-7B \cite{wang2024qwen2} &  0.737/0.514  &  0.656/0.916 &  0.666/0.705  &  0.888/0.699  &  0.827/0.765  & 0.918/0.553  & 0.782/0.692 \\

     LLava 1.6-7B \cite{liu2024llava} &  0.768/0.514 &  0.653/0.918 & 0.683/0.697 & 0.758/0.101 & 0.674/0.762 & 0.545/0.425 & 0.680/0.569 \\

     \rowcolor{gray!20} AGHI-Assessor (Ours) & \bf{0.961/0.709}  & \bf{0.991/0.923} & \bf{0.985/0.782} & \bf{0.955/0.896}  & \bf{0.914/0.780} &   \bf{0.970/0.695} & \bf{0.962/0.797} \\

    \bottomrule[1pt]
  \end{tabular}
  \label{benchmark_res}
\end{table*}

\subsection{Experimental Setup}

\noindent\textbf{Comparison Algorithms.} To evaluate the performance of our AGHI-Assessor, we select the state-of-the-art evaluation metrics for comparison, which could be divided into multiple groups: (1) Traditional no-reference IQA methods, including NIQE \cite{mittal2012making} and BRISQUE \cite{mittal2011blind}. (2) Vision language pre-training models, including BLIPScore \cite{li2023blip}, CLIPScore \cite{radford2021learning}, Aesthetic Score \cite{schuhmann2022laion} and PickScore \cite{kirstain2023pick}. (3) Deep learning-based no-reference IQA methods, including DBCNN \cite{zhang2018blind}, HyperIQA \cite{su2020blindly}, MUSIQ \cite{ke2021musiq}, MANIQA \cite{yang2022maniqa}, UNIQUE \cite{zhang2021uncertainty}, DN-PIQA \cite{sun2024dual}, DSL-FIQA \cite{chen2024dsl}. (4) deep learning-based AI-generated IQA methods, including MA-AGIQA \cite{wang2024large}, IPCE \cite{peng2024aigc}. (5) LMM question answering models, we select Deeepseek-VL-7B \cite{lu2024deepseek}, Intern-VL 2-8B \cite{team2024internvl2}, Qwen-VL 2.5-7B \cite{bai2025qwen2} and LLaVA-1.6-7B \cite{liu2024llava} to evaluate the understanding performance of AI-generated human images.

\begin{figure}[ht]
\centering
\centerline{\includegraphics[width=0.99\linewidth]{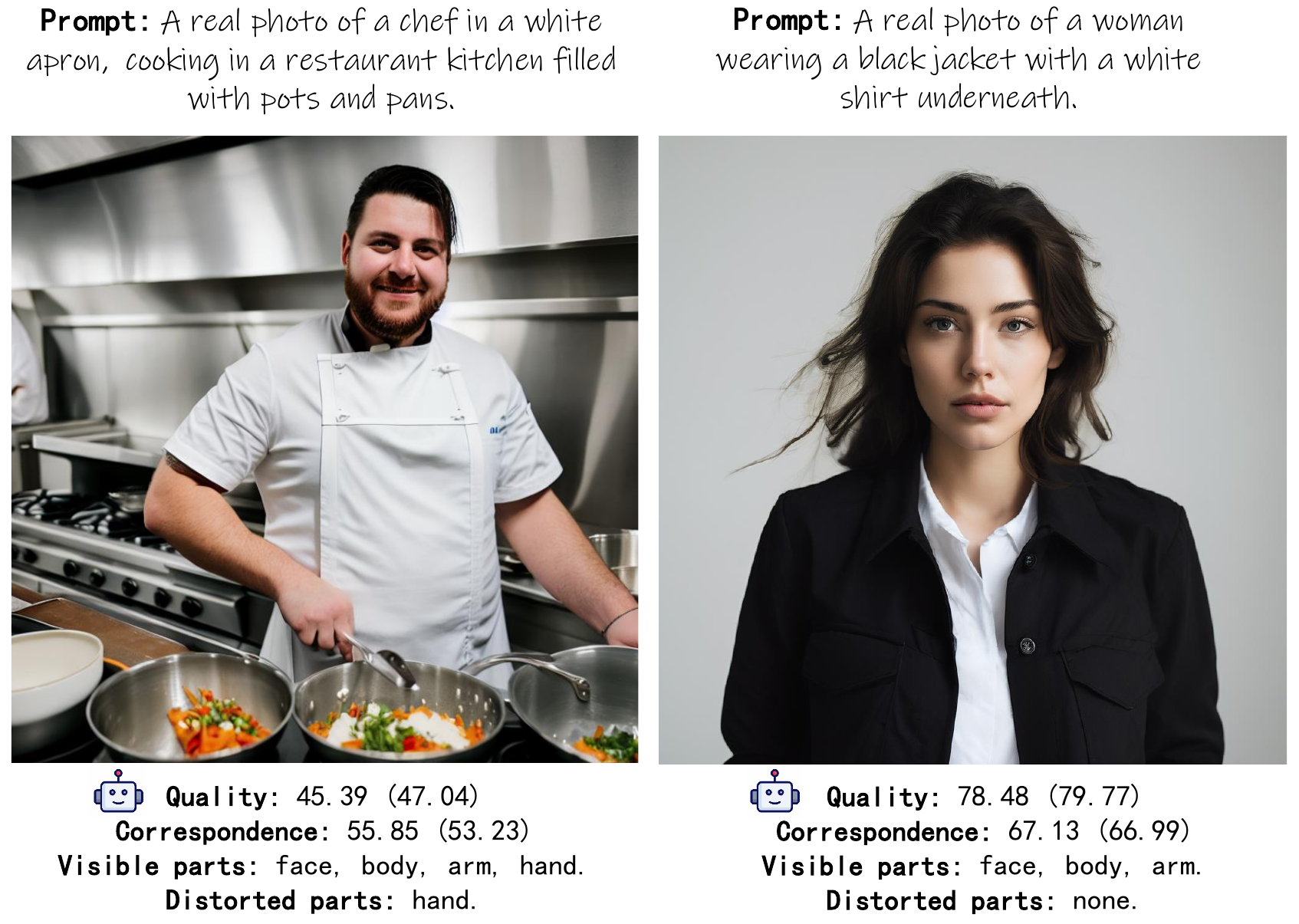}}
\caption{ \textbf{Qualitative results} of our predicted results. Our model can precisely produce multi-dimensional scores and identifying the visible and distorted human parts.}
\label{qual_res}
\end{figure}

\begin{table*}
\renewcommand\arraystretch{1.2}
  \caption{Ablation study of different components in our proposed method. ``Human-center", ``AGI Feat", ``Text Feat" denote adding cropped human-centered image, AI-generated image quality-aware feature, and text feature.}
  \resizebox{0.74\textwidth}{!}{
  \begin{tabular}{ccccc| ccc ccc}
    \toprule
    & \multicolumn{4}{c}{Feature \& Strategy}                                  & \multicolumn{3}{c}{Perceptual Quality}                   & \multicolumn{3}{c}{Text-Image Correspondence} \\
    \cmidrule(r){2-5} \cmidrule(r){6-8} \cmidrule(r){9-11} 
    \multicolumn{1}{c}{No.}   & Human-centered  & AGI Feat   & Text Feat & LoRA  & SRCC     & PLCC     & KRCC          & SRCC       & PLCC    & KRCC        \\
    \multicolumn{1}{c}{(1)}   &    &   &    &  \ding{52}  & 0.8218 &  0.8267 & 0.6339 &  0.7754 &  0.7775  & 0.5803 \\

     \multicolumn{1}{c}{(2)}          &    &      \ding{52}     &            &     \ding{52}     &  0.8318 & 0.8340 & 0.6443 &  0.7728 & 0.7763 & 0.5810   \\

     \multicolumn{1}{c}{(3)}      &  \ding{52}   &      \ding{52}     &            &     \ding{52}     &   0.8385 & 0.8492 & 0.6574  & 0.7806 & 0.7850 & 0.5887 \\

\multicolumn{1}{c}{(4)}      &  \ding{52}   &      \ding{52}     &    \ding{52}          &        &  0.8143 & 0.8280 & 0.6214 &  0.7528 & 0.7617 & 0.5631 \\

     \rowcolor{gray!20} \multicolumn{1}{c}{(5)}      &  \ding{52}  &   \ding{52}  &     \ding{52}  &  \ding{52}  &   \bf{0.8447} & \bf{0.8504} & \bf{0.6625} & \bf{0.7818} & \bf{0.7864} & \bf{0.5907}   \\
     
    \bottomrule
  \end{tabular}\label{ablation}
   }
  \centering
\end{table*}

\noindent\textbf{Evaluating Settings.} We evaluate AGHI-Assessor on both our AGHI-QA dataset and AGIQA-3K \cite{li2023agiqa} dataset. During evaluation, we split the data set into training set and testing set with a ratio of 0.8 and 0.2. We randomly split the data set five times and report the average results. Spearman rank correlation coefficient (SRCC), Pearson linear correlation coefficient (PLCC) and Kendall rank-order correlation coefficient (KRCC) are utilized for evaluating the scoring ability of each model. For vision language pre-training methods, we utilize their pre-trained models for zero-shot inference. For deep learning-based methods, all methods are trained in our data set with the same training and testing set. Considering that our AGHI-Assessor is specifically designed for AI-generated human images, we remove the cropped human-centered images for the evaluation on the AGIQA-3K dataset.

\begin{table}[tp]
\renewcommand\arraystretch{1.1}
\caption{Performance comparisons on the AGIQA-3K database. The best results are bolded in the table.}
   \footnotesize
			\label{tab:3k_results}
			\centering
				\begin{tabular}{llccc}
					\toprule
					\multicolumn{1}{l}{\textbf{Dimension} }  &\textbf{Method}                      & \textbf{SRCC} & \textbf{PLCC}  & \textbf{KRCC}      \\   \midrule
            \multirow{17}{*}{Quality}&CEIQ \cite{yan2019no}                    & 0.3228  & 0.4166 & 0.2220 \\
     &DSIQA \cite{narvekar2011no}                   & 0.4955  & 0.5488 & 0.3403  \\
     &NIQE \cite{mittal2012making}                    & 0.5623  & 0.5171 & 0.3876 \\
     &FID \cite{heusel2017gans}                     & 0.1733 & 0.1860  & 0.1158  \\
    &ICS \cite{salimans2016improved}                     & 0.0931  & 0.0964 & 0.0626  \\
    &KID \cite{binkowski2018demystifying}                     & 0.1023  & 0.0786 & 0.0692  \\ 
    & GMLF \cite{xue2014blind}    & 0.6987  & 0.8181 & 0.5119  \\
    & Higrade \cite{kundu2017large}    & 0.6171  & 0.7056 & 0.4410  \\ 
    & DBCNN \cite{zhang2018blind}      & 0.8207  & 0.8759 & 0.6336\\
    & CNNIQA \cite{kang2014convolutional}                  & 0.7478  & 0.8469 & 0.5580\\
    & HyperNet \cite{su2020blindly}                & 0.8355 & 0.8903  & 0.6488\\ 
    & CLIP-IQA+~\cite{wang2023exploring} & 0.8428 & 0.8879 & 0.6556\\
    & CLIP-AGIQA~\cite{fu2024vision} & {0.8747} & {0.9190} & {0.6976} \\
    & IPCE~\cite{peng2024aigc} & {0.8841} & {0.9246} & {0.7091} \\

     \cdashline{2-5}
     \rowcolor{gray!20}&\textbf{AGHI-Assessor (Ours)}      & \bf{0.9038} & \bf{0.9322}  & \bf{0.7411} \\
     \hline
     \multirow{7}{*}{Alignment}&
     CLIP \cite{radford2021learning}                                          & 0.5972 & 0.6839  & 0.4591 \\
					
	&HPS \cite{wu2023human}                                          & 0.6349  & 0.7000 & 0.4580  \\
	&PickScore \cite{kirstain2023pick}                                    & 0.6977 & 0.7633  & 0.5069\\
     &ImageReward \cite{xu2024imagereward}                                  & 0.7298& 0.7862 & 0.5390\\
     &StairReward  \cite{li2023agiqa}                                   & 0.7472  & 0.8529 & 0.5554
    \\  
    \cdashline{2-5}
    \rowcolor{gray!20} &\textbf{AGHI-Assessor (Ours)}       & {\bf{0.8479}} & {\bf{0.8830}}  & {\bf{0.6899}} \\ 

\hline
    
\end{tabular}
\end{table}

\subsection{Implementation Details}

For the image encoder for both original AI-generated human images and cropped human images, we select the pre-trained InternVIT \cite{chen2024internvl} as the visual backbone. We resize each image to 448 resolutions and then feed it into images encoder. For AIGC image encoder, we select the PickScore \cite{kirstain2023pick} model, which is learned from a large dataset of general AI-generated image quality assessment. As for text encoder, we select the BLIP text encoder to extract the overall text feature for each prompt sentence. The LLM utilized in our AGHI-Assessor is InternLM 2.5-8B \cite{chen2024expanding}, with the language token dimension being 4096. During training, we freeze the image encoder, text encoder, AIGC image encoder and LLM. The rank parameters of LoRA for LLM and visual encoder are set to 16 and 8 respectively. The model is trained on 2 NVIDIA RTX A6000 GPUs and flash attention is used to save the GPU memory. We train the network for 5 epochs with a total batch size of 16.

\subsection{Performance and Analysis }

The results are listed in Tab. \ref{benchmark_res}. From the table, we can observe that traditional NR IQA models show poor performance in evaluating both the perceptual quality and the text-image correspondence of AI-generated human images. The reason may be that traditional features are not suitable for evaluating the semantic distortions of human body parts in AI-generated human images. Compared to traditional NR IQA models, deep learning-based IQA and AI-generated IQA models  generally exhibit better performance across all two scores. Among all IQA models, DSL-FIQA \cite{chen2024dsl} achieves the best performance on predicting perceptual quality and the second best performance on predicting text-image correspondence, demonstrating excellent prediction ability and strong generalization capabilities of modeling the quality feature from AI-generated human images. IPCE \cite{peng2024aigc} achieves the best performance on predicting text-image correspondence scores because it uses a CLIP-based method which can effectively model the relationship between input texts and generated images. Besides, we find that current vision language pre-training methods also show poor performance when predicting all scores. PickScore \cite{kirstain2023pick} achieves the best performance because it is trained on a very large-scale general AIGC image dataset containing about 500k images, so it can generally capture the quality of AI-generated human images. Notably, our AGHI-Assessor performs best compared to all other methods, demonstrating the leading advantage of the LMM-based quality assessment method and our kindly design framework. We also provide some prediction results on our dataset in Fig. \ref{qual_res}, demonstrating the great performance of AGHI-Assessor.

To demonstrate the generalization of our method, we also report quality assessment experimental results on the AGIQA3K \cite{li2023agiqa} dataset, which is a general AI-generated image QA dataset. From the results shown in Table \ref{tab:3k_results}, our AGHI-Assessor demonstrates excellent scoring performance on both quality dimensions, showing the potential of our proposed method.

\subsection{Identifying Visible and Distorted Human Body Parts}

Generating a simple quality score is not sufficient for a comprehensive evaluation. With the help of LMM models, we present a new task of generating descriptions of presence and distortions of human body parts for each AI-generated human image. We benchmark the performance of current image LMM models for this task in Table \ref{benchmark_res}. From the results, we can observe that InternVL 2 performs the best on identifying the visibility of human body part. However, it struggles to accurately detect whether those body parts exhibit semantic distortions. The Qwen VL 2.5-7B performs the best considering the overall performance on both questions. We find that identifying visible body parts is a relatively easier task for current LMMs compared to detecting distorted parts. Concretely, detecting distortions of hand, face and foot proves to be more challenging, while identifying the distortions in the body, leg and arm is comparatively easier. The reason is that hand and face are quite complex for LMM to checking the detailed semantic distortions such as missing a finger or one unrealistic eye. Overall, our AGHI-Assessor demonstrates the best performance in both visible parts detection and distortion identification tasks, with a large performance improvement.

\subsection{Ablation Study}

Extensive ablation studies are conducted to demonstrate the importance of each component in AGHI-Assessor. The results are summarized in Table \ref{ablation}.

\subsubsection{Effectiveness of Cropping Human-Centered Image} We validate the importance of the human-centered image cropping strategy. As shown in the third row of Table \ref{ablation}, introducing the cropped human image significantly improves the final performance. The validate the assumption that participants are prone to focus on the area of human subject in each AI-generated human image.

\subsubsection{Effectiveness of AGI-aware Quality Encoder} To assess the impact of the AGI-aware quality encoder, we conduct ablation studies to verify the effectiveness of the PickScore features for general AI-generated images. As evidenced by the first two rows of Table \ref{ablation}, incorporating PickScore features \cite{kirstain2023pick} results in a marked performance gain in our data set compared to the baseline. This underscores the value of integrating motion features into the proposed framework.

\subsubsection{Effectiveness of Text Encoder} To evaluate the contribution of the text encoder, we perform ablation studies to assess the impact of incorporating detailed sentence features from prompts. As shown in row of Table \ref{ablation}, the inclusion of text features leads to a notable performance improvement on the AGHI-QA dataset compared to the baseline. This demonstrates the critical role of text features in enhancing the overall effectiveness of our AGHI-Assessor model.

\subsubsection{Effectiveness of LoRA Adaptation} Finally, we evaluate the effectiveness of applying LoRA (Low-Rank Adaptation) on AGHI-Assessor by analyzing its impact on model performance, as illustrated in the second row from the bottom of Table \ref{ablation}. The result shows that applying LoRA to both the image encoder ($R_{lora}=8$ for vision encoder) and the large language model ($R_{lora}=16$ for llm) largely improves the model's performance.

\section{Conclusion}

In this work, we conduct the first comprehensive quality assessment study specifically for AGHIs. In particular, we introduce AGHI-QA, the first large-scale benchmark specifically designed for quality assessment of AGHIs. The dataset comprises $4,000$ images generated from $400$ carefully crafted text prompts using $10$ state-of-the-art T2I models. We conduct a systematic subjective study to collect multidimensional annotations, including perceptual quality scores, text-image correspondence scores, visible and distorted body part labels. Based on AGHI-QA, we evaluate the strengths and weaknesses of current T2I methods in generating human images from multiple dimensions. Furthermore, we propose AGHI-Assessor, a novel quality metric that integrates the large multimodal model (LMM) with domain-specific human features for precise quality prediction and identification of visible and distorted body parts in AGHIs. Extensive experimental results demonstrate that AGHI-Assessor showcases state-of-the-art performance, significantly outperforming existing IQA methods in multidimensional quality assessment and surpassing leading LMMs in detecting structural distortions in AGHIs.

\bibliographystyle{IEEEtran}
\bibliography{refs}

\vfill

\end{document}